\documentclass[final]{cvpr}

\usepackage{times}
\usepackage{epsfig}
\usepackage{graphicx}
\usepackage{amsmath}
\usepackage{amssymb}

\usepackage[utf8]{inputenc} 
\usepackage[T1]{fontenc}    
\usepackage{url}            
\usepackage{booktabs}       
\usepackage{amsfonts}       
\usepackage{nicefrac}       
\usepackage{microtype}      

\usepackage{subfigure}
\usepackage{caption}

\usepackage[abs]{overpic}
\usepackage{wrapfig}

\usepackage{graphicx}

\usepackage{xcolor}
\usepackage{textcomp} 
\usepackage{mdwlist}

\newcommand{\trsp}{\mathsf{T}}
\newcommand{\real}{\mathbb{R}}
\newcommand{\height}{H}
\newcommand{\width}{W}

\usepackage[pagebackref=true,breaklinks=true,colorlinks,bookmarks=false]{hyperref}

\begin{document}
\title{Motion Representations for Articulated Animation}
\author{Aliaksandr Siarohin$^{1*}$, Oliver J. Woodford$^*$, Jian Ren$^2$, Menglei Chai$^2$ and Sergey Tulyakov$^2$\\
$^1$DISI, University of Trento, Italy, $^2$Snap Inc., Santa Monica, CA
\\
{\tt\small aliaksandr.siarohin@unitn.it},
{\tt\small \{jren,mchai,stulyakov\}@snap.com},
{\scriptsize $^*$Work done while at Snap Inc.}}
\maketitle

\begin{abstract}
%

We propose novel motion representations for animating articulated objects consisting of distinct parts. 
In a completely unsupervised manner, our method identifies object parts, tracks them in a driving video, and infers their motions by considering their principal axes. 
In contrast to the previous keypoint-based works, our method extracts meaningful and consistent \emph{regions}, describing locations, shape, and pose. The regions correspond to semantically relevant and distinct object parts, that are more easily detected in frames of the driving video. To force decoupling of foreground from background, we model non-object related global motion with an additional affine transformation. 
%
To facilitate animation and prevent the leakage of the shape of the driving object, we disentangle shape and pose of objects in the region space.
Our model\footnote{Our source code is publicly available at \href{https://github.com/snap-research/articulated-animation}{https://github.com/snap-research/articulated-animation}.} can animate a variety of objects, surpassing previous methods by a large margin on existing benchmarks. We present a challenging new benchmark with high-resolution videos and show that the improvement is particularly pronounced when articulated objects are considered, reaching 96.6\% user preference vs.\ the state of the art. 
\end{abstract}

\vspace{-0.4cm}
\section{Introduction}
\label{sec:intro}
\begin{figure}[h]
    \centering
    \includegraphics[width=\linewidth]{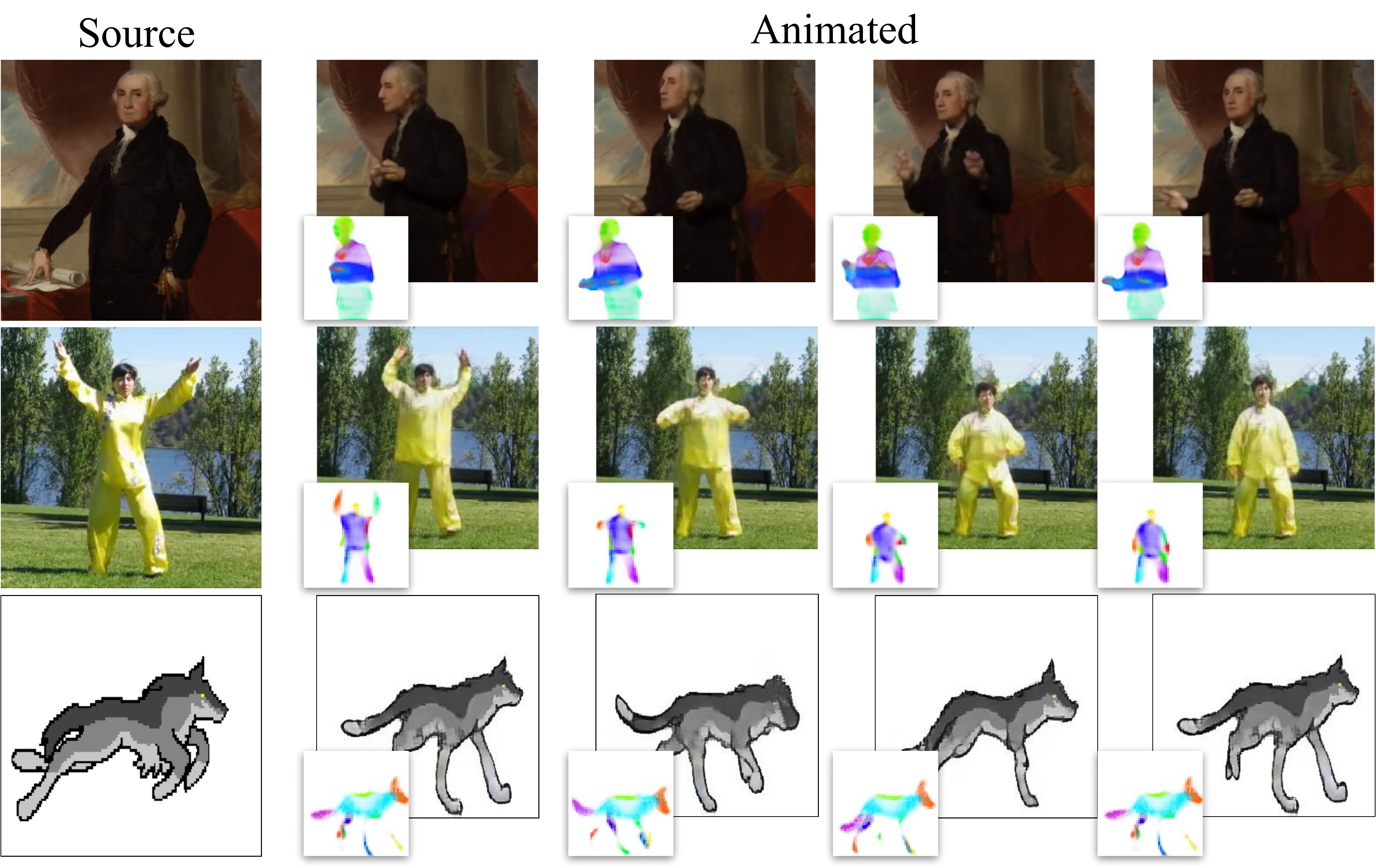}
    \caption{Our method animates still source images via unsupervised region detection (inset).}
    \label{fig:teaser}
    \vspace{-0.4cm}
\end{figure}

\vspace{-0.2cm}
Animation---bringing static objects to life---has broad applications across education and entertainment. Animated characters and objects, such as those in Fig.~\ref{fig:teaser}, increase the creativity and appeal of content, improve the clarity of material through storytelling, and enhance user experiences\footnote{Visit our project \href{https://snap-research.github.io/articulated-animation}{website} for more qualitative samples.}.  

Until very recently, animation techniques necessary for achieving such results required a trained professional, specialized hardware, software, and a great deal of effort. Quality results generally still do, but vision and graphics communities have attempted to address some of these limitations by training data-driven methods~\cite{wang2018video, chan2019everybody, ren2020human, geng20193d,gafni2019vid2game} on object classes for which prior knowledge of object shape and pose can be learned. This, however, requires ground truth pose and shape data to be available during training. 

Recent works have sought to avoid the need for ground truth data through \emph{unsupervised} motion transfer~\cite{wiles2018x2face, fomm,siarohin2018animating}. Significant progress has been made on several key challenges, including training using image reconstruction as a loss~\cite{wiles2018x2face, fomm,siarohin2018animating}, and disentangling motion from appearance~\cite{lorenz2019unsupervised}. This has created the potential to animate a broader range of object categories, without any domain knowledge or labelled data, requiring only videos of objects in motion during training~\cite{fomm}. However, two key problems remain open. The first is how to represent the parts of an articulated or non-rigid moving object, including their shapes and poses. The second is given the object parts, how to animate them using the sequence of motions in a driving video.
    
Initial attempts used  end-to-end frameworks~\cite{wiles2018x2face, siarohin2018animating} to first extract unsupervised keypoints~\cite{lorenz2019unsupervised,kim2019unsupervised}, then warp a feature embedding of a source image to align its keypoints with those of a driving video. Follow on work~\cite{fomm} further modelled the motion around each keypoint with local, affine transformations, and introduced a generation module that both composites warped source image regions and inpaints occluded regions, to render the final image. This enabled a variety of creative applications,\!\!\footnote{E.g.\ a \href{https://www.youtube.com/watch?v=Ac6kDNMQK3c}{music video} in which images are animated using prior work~\cite{fomm}.} for example needing only one source face image to generate a near photo-realistic animation, driven by a video of a different face.

However, 
the resulting unsupervised keypoints are detected on the boundary of the objects. While points on edges are easier to identify, tracking such keypoints between frames is problematic, as any point on the boundary is a valid candidate, making it hard to establish correspondences between frames. A further problem is that the unsupervised keypoints do not correspond to semantically meaningful object parts, representing location and direction, but not shape. Due to this limitation, animating articulated objects, such as bodies, remains challenging. Furthermore, these methods assume static backgrounds, i.e.\ no camera motion, leading to leakage of background motion information into one or several of the detected keypoints. Finally, absolute motion transfer, as in~\cite{fomm}, transfers the shape of the driving object into the generated sequence, decreasing the fidelity of the source identity.
These remaining deficiencies limit the scope of previous works~\cite{fomm,siarohin2018animating} to more trivial object categories and motions, especially when objects are articulated.

This work introduces three contributions to address these challenges. First, we redefine the underlying motion representation, using \emph{regions} from which first-order motion is \emph{measured}, rather than regressed. This enables improved convergence, more stable, robust object and motion representations, and also empirically captures the shape of the underpinning object parts, leading to better motion segmentation. This motion representation is inspired by Hu moments~\cite{davis1997representation}.
Fig.~\ref{fig:keypoints-qualitative}(a) contains several examples of region vs.\ keypoint-based motion representation.

Secondly, we explicitly model background or camera motion between training frames by predicting the parameters of a global, affine transformation explaining non-object related motions. This enables the model to focus solely on the foreground object, making the identified points more stable, and further improves convergence.
Finally, to prevent shape transfer and improve animation, we disentangle the shape and pose of objects in the space of unsupervised regions. Our framework is self-supervised, does not require any labels, and is optimized using reconstruction losses.

These contributions further improve unsupervised motion transfer methods, resulting in higher fidelity animation of articulated objects in particular. To create a more challenging benchmark for such objects, we present a newly collected dataset of TED talk speakers. Our framework scales better in the number of unsupervised regions, resulting in more detailed motion. Our method outperforms previous unsupervised animation methods on a variety of datasets, including talking faces, taichi videos and animated pixel art being preferred by 96.6\% of independent raters when compared with the state of the art~\cite{fomm} on our most challenging benchmark.
\vspace{-0.2cm}
\section{Related work}
\vspace{-0.2cm}
Image animation methods can be separated into supervised, which require knowledge about the animated object during training, and unsupervised, which do not. Such knowledge typically includes landmarks~\cite{cao2014displaced, zakharov2019few, Qian_2019_ICCV,ha2019marionette}, semantic segmentations~\cite{Nirkin_2019_ICCV}, and parametric 3D models~\cite{geng20193d, thies2016face2face, deng2020disentangled, nagano2018pagan,liu2019liquid}. As a result, supervised methods are limited to a small number of object categories for which a lot of labelled data is available, such as faces and human bodies.
Early face reenactment work~\cite{thies2016face2face} fitted a 3D morphable model to an image, animating and rendering it back using graphical techniques. Further works used neural networks to get higher quality rendering~\cite{kim2018deep, wang2018every}, sometimes requiring multiple images per identity~\cite{geng20193d, pumarola2018ganimation}. A body of works treats animation as an image-to-image~\cite{siarohin2018deformable} or video-to-video~\cite{wang2018video, chan2019everybody,ren2020human} translation problem. Apart from some exceptions~\cite{wang2019few}, these works further constrain the problem to animating a single instance of an object, such as a single face~\cite{kim2018deep,bansal2018recycle} or a single human body~\cite{chan2019everybody, ren2020human, wang2018video}, requiring retraining~\cite{bansal2018recycle,chan2019everybody,ren2020human} or fine-tuning~\cite{zakharov2019few} for each new instance. Despite promising results, generalizing these methods beyond a limited range of object categories remains challenging. Additionally, they tend to transfer not only the motion but also the shape of the driving object~\cite{kim2018deep, zakharov2019few}.

Unsupervised methods address some of these limitations. They do not require any labelled data regarding the shape or landmarks of the animated object. 
Video-generation-based animation methods predict future frames of a video, given the first frame and an animation class label, such as ``make a happy face'', ``do jumping jack'', or ``play golf''~\cite{tulyakov2017mocogan, saito2017temporal, clark2019adversarial}. Recently Menapace \etal~\cite{menapace2021playable} introduce playable video generation, where action could be selected at each timestamp. A further group of works re-target animation from a driving video to a source frame.
X2Face~\cite{wiles2018x2face} builds a canonical representation of an input face, and generates a warp field conditioned on the driving video. 
Monkey-Net~\cite{siarohin2018animating} learns a set of unsupervised keypoints to generate animations. Follow-up work substantially improves the quality of animation by considering a first order motion model (FOMM)~\cite{fomm} for each keypoint, represented by regressing a local, affine transformation. Both of these works apply to a wider range of objects including faces, bodies, robots, and pixel art animations. Empirically, these methods extract keypoints on the boundary of the animated objects. Articulated objects such as human bodies are therefore challenging, as internal motion, for example, an arm moving across the body, is not well modeled, producing unconvincing animations.

This work presents an unsupervised method. We argue that the limitations of previous such methods in animating articulated objects is due to an inability of their internal representations to capture complete object parts, their shape and pose. X2Face~\cite{wiles2018x2face} assumes an object can be represented with a single RGB texture, while other methods find keypoints on edges~\cite{siarohin2018animating,fomm}.
Our new region motion representation resembles the construction of a motion history image whose shape is analyzed using principal components (namely Hu moments~\cite{davis1997representation}). In \cite{davis1997representation} the authors construct motion descriptors by computing temporal image differences, aggregate them into motion history images, and use Hu moments to build a motion recognition system. In this manner, blob statistics are used to discriminate between different actions.

\vspace{-0.2cm}
\section{Method}
\label{sec:method}
\vspace{-0.2cm}
We propose three contributions over FOMM~\cite{fomm}, namely our PCA-based motion estimation (Sec.~\ref{sec:regions}), background motion representation (Sec.~\ref{sec:bg}) and animation via disentanglement (Sec.~\ref{sec:animation}). To make the manuscript self-contained, we first give the necessary technical background on the original first order motion representation~\cite{fomm} (Sec.~\ref{sec:fomm}).

\begin{figure*}[t]
    \centering
    \includegraphics[width=0.94\linewidth]{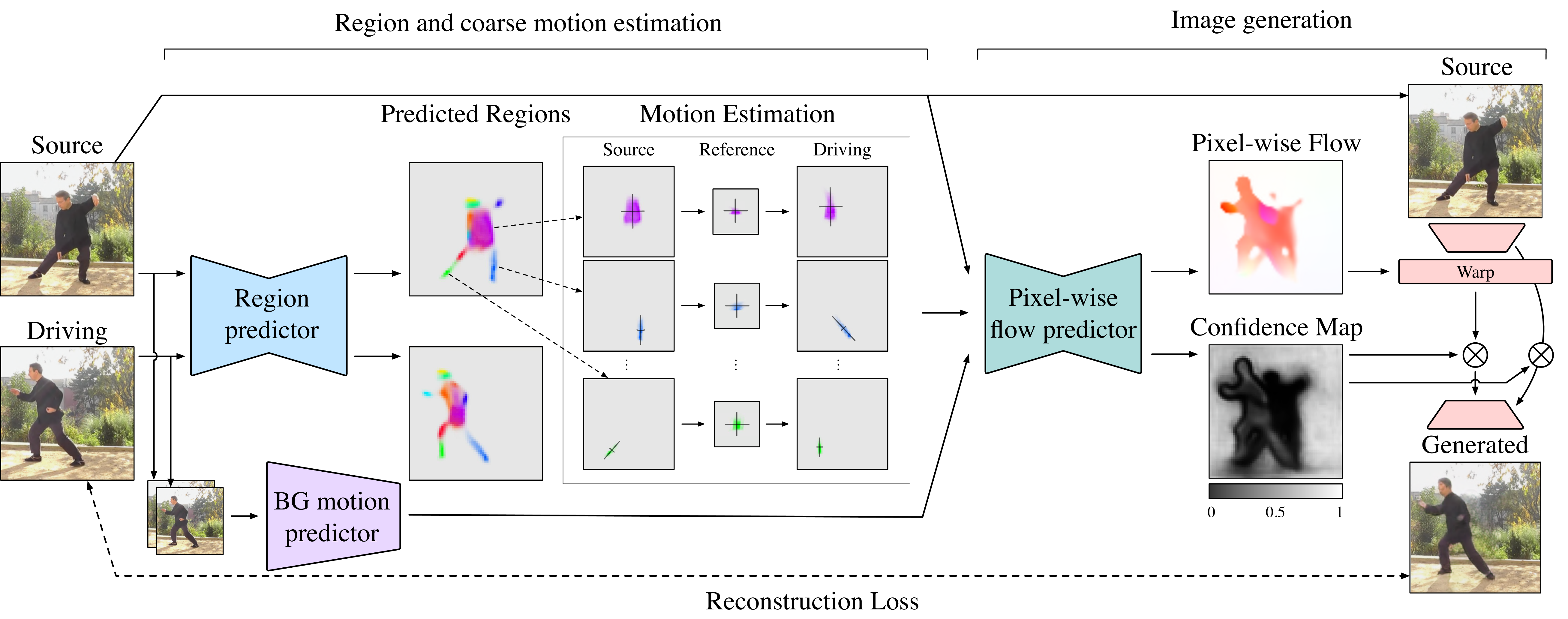}
    \vspace{-0.2cm}
    \caption{\textbf{Overview of our model.} The region predictor returns heatmaps for each part in the source and the driving images. We then compute principal axes of each heatmap, to transform each region from the source to the driving frame through a whitened reference frame. 
    Region and background transformations are combined by the pixel-wise flow prediction network. The target image is generated by warping the source image in a feature space using the pixel-wise flow, and inpainting newly introduced regions, as indicated by the confidence map.}
    \label{fig:framework}
    \vspace{-0.4cm}
\end{figure*}

\newcommand{\nregions}{K}
\newcommand{\heatmap}{\mathbf{M}}
\newcommand{\heatmapval}{m}
\newcommand{\imcoord}{z}
\newcommand{\imcoordset}{\mathcal{Z}}
\newcommand{\affine}{A}
\newcommand{\imX}{\mathbf{X}}
\newcommand{\sourceframe}{\mathbf{S}}
\newcommand{\drivingframe}{\mathbf{D}}
\newcommand{\refframe}{\mathbf{R}}
\newcommand{\opticalflow}{\mathbf{O}}
\newcommand{\assignment}{\mathbf{W}}
\newcommand{\confidence}{\mathbf{C}}
\newcommand{\DtoS}{{\sourceframe \leftarrow \drivingframe}}
\newcommand{\RtoX}{{\imX \leftarrow \refframe}}
\newcommand{\RtoS}{{\sourceframe \leftarrow \refframe}}
\newcommand{\RtoD}{{\drivingframe \leftarrow \refframe}}
\newcommand{\covar}{\Sigma}
\newcommand{\mean}{\mu}
\renewcommand{\angle}{\theta}
\newcommand{\U}{U}
\renewcommand{\S}{S}
\newcommand{\V}{V}
\newcommand{\rot}{R}
\renewcommand{\covar}{\Sigma}

\vspace{-0.1cm}
\subsection{First Order Motion Model}
\label{sec:fomm}
\vspace{-0.1cm}
FOMM~\cite{fomm} consists of the two main parts: motion estimation and image generation, where motion estimation further contains coarse motion estimation and dense motion prediction. Coarse motion is modelled as sparse motions between separate object parts, while dense motion produces an optical flow along with the confidence map for the entire image. We denote by $\sourceframe$ and $\drivingframe$ the source and the driving frames extracted from the same video respectively.

The first step is to estimate coarse motions from $\sourceframe$ and $\drivingframe$. Motions for each object part are represented by affine transformations, $\affine^k_\RtoX \in \mathcal{R}^{2\times3}$, to an abstract, common reference frame, $\refframe$; $\imX$ is either $\sourceframe$ or $\drivingframe$. Motions are estimated for $\nregions$ distinct parts. An encoder-decoder \emph{keypoint} predictor network outputs $\nregions$ heatmaps, $\heatmap^1,..,\heatmap^\nregions$ for the input image, followed by softmax, s.t.~$\heatmap^k~\in[0,1]^{\height \times \width}$, where $\height$ and $\width$ are the height and width of the image respectively, and $\sum_{\imcoord\in\imcoordset}\heatmap^k(\imcoord) = 1$, where $\imcoord$ is a pixel location (x, y coordinates) in the image, the set of all pixel locations being $\imcoordset$, and $\heatmap^k(\imcoord)$ is the $k$-th heatmap weight at pixel $\imcoord$. Thus, the translation component of the affine transformation (which is the last column of $\affine^k_\RtoX$) can be estimated using softargmax:
\begin{align}
\label{eq:shif}
\mean^k &= \sum_{\imcoord\in\imcoordset} \heatmap^k(\imcoord) \imcoord.
\end{align}
In FOMM~\cite{fomm} the remaining affine parameters are regressed per pixel and form 4 additional channels  $P^k_{ij}~\in\mathcal{R}^{\height \times \width}$, where $i~\in\{0,1\},j~\in\{0,1\}$ indexes of the affine matrix $\affine^k_\RtoX$. The latter is estimated using weighted pooling:
\begin{equation}
\affine^k_\RtoX[i, j] = \sum_{\imcoord\in\imcoordset} \heatmap^k(\imcoord) P^k_{ij}(z).
\end{equation}
We refer to this way of computing motion as \emph{regression-based}, where affine parameters are \emph{predicted} by a network and pooled to compute $\affine^k_\RtoX$.
Motion between $\drivingframe$ and $\sourceframe$ for part $k$ is then computed via the common reference frame:
\begin{align}
    \label{eq:driving-to-source}
    \affine_\DtoS^k &= \affine_\RtoS^k \begin{bmatrix}\affine_\RtoD^k\\ 0~~0~~1\end{bmatrix}^{-1}.
\end{align}
Given the coarse motion, the next step is to predict the optical flow and the confidence map. 
Since the transformations between $\drivingframe$ and $\sourceframe$ are known for each part, the task is to combine them to obtain a single optical flow field. To this end, the flow predictor selects the appropriate coarse transformation for each pixel, via a weighted sum of coarse motions. Formally, $K+1$ assignment maps are output, one per region, plus background, which FOMM~\cite{fomm} assumes is \emph{motionless}, and softargmax is applied pixelwise across them, s.t.\ $\assignment^k~\in[0,1]^{\height \times \width}$, $\assignment^0$ corresponds to the background, and $\sum_{k=0}^K\assignment^k(z) = 1,~\forall z$. Optical flow per pixel, $\opticalflow(z)\in \real^2$, is then computed as:
\begin{equation}
    \opticalflow(z) = \assignment^0(z)z + \sum_{k=1}^K \assignment^k(z) \affine^k_\DtoS \begin{bmatrix} z \\ 1\end{bmatrix}.
    \label{eq:flow}
\end{equation}
With such a model, animation becomes challenging when there is even slight background motion. The model automatically adapts by assigning several of the available keypoints to model background as shown in the first row of Fig.~\ref{fig:keypoints-qualitative}(a). 

A confidence map, $\confidence~\in[0,1]^{\height \times \width}$, is also predicted using the same network, to handle parts missing in the source image. Finally $\sourceframe$ is passed through an encoder, followed by warping the resulting feature map using the optical flow (Eq.~\ref{eq:flow}) and multiplied by the confidence map. A decoder then reconstructs the driving image $\drivingframe$.

At test time FOMM~\cite{fomm} has two modes of animating $\sourceframe$: \emph{standard} and \emph{relative}. In both cases the input is the source image $\sourceframe$ and the driving video $\drivingframe_1, \drivingframe_2,.., \drivingframe_t,.., \drivingframe_T$. In the \emph{standard} animation the motion between source and driving is computed frame-by-frame using Eq.~\eqref{eq:driving-to-source}.
For \emph{relative} animation, in order to generate a frame $t$ the motion between $\drivingframe_1$ and $\drivingframe_t$ is computed first and then applied to $\sourceframe$. Both of these modes are problematic when the object in question is articulated, as we show in Sec.~\ref{sec:animation}.


\vspace{-0.2cm}
\subsection{PCA-based motion estimation}
\label{sec:regions}
\begin{figure*}[t]
    \centering
    \includegraphics[width=\linewidth]{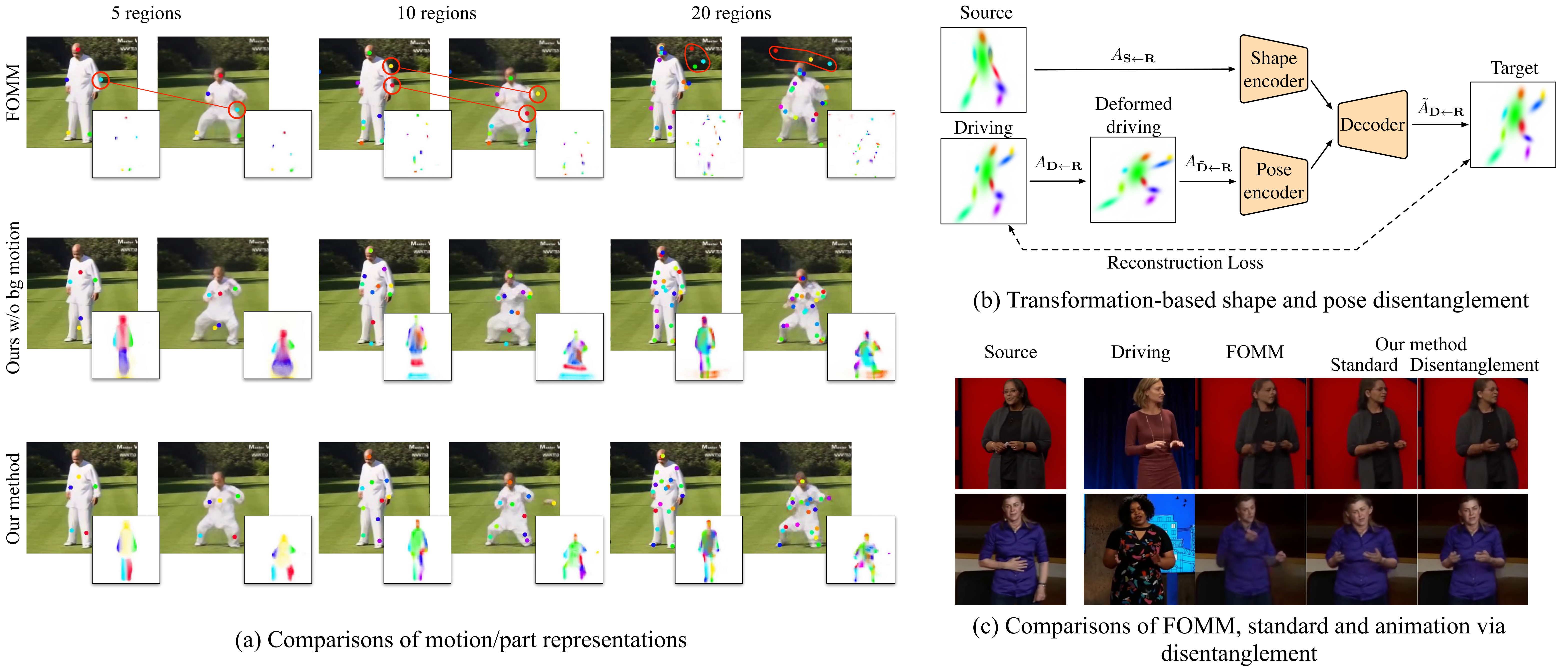}
    \vspace{-0.7cm}
    \caption{\textbf{Motion representation and disentanglement.} (a) A comparison of part regions estimated by regression-based (FOMM~\cite{fomm}) and PCA-based (our method) motion representations. Each row shows a generated sequence along with the detected keypoints and region heatmaps (inset).
    (b) Framework for animating motion driving frame whilst retaining shape from the source frame, by disentangling shape and pose from motion representations. (c) Qualitative animation results of articulated objects, using FOMM~\cite{fomm}, and our method using standard and disentangled motion transfer (sec.~\ref{sec:animation}).
    }
    \label{fig:keypoints-qualitative}
\end{figure*}

\vspace{-0.2cm}
Accurate motion estimation is the main requirement for high-quality image animation. As mentioned previously, FOMM regresses the affine parameters. This requires higher capacity networks, and generalizes poorly (see Sec.~\ref{sec:toy}). We propose a different motion representation: \emph{all} motions are \emph{measured} directly from the heatmap $\heatmap^k$. We compute the translation as before, while in-plane rotation and scaling in x- and y-directions are computed via a principal component analysis (PCA) of the heatmap $\heatmap^k$. 
Formally, the transformation $\affine^k_\RtoX \in \real^{2\times 3}$, of the $k^\textrm{th}$ region from the reference frame to the image is computed as:
\begin{align}
\mean^k &= \sum_{\imcoord\in\imcoordset} \heatmap^k(\imcoord) \imcoord, \\
\label{eq:pca}
\U^k \S^k {\V^k} &= \sum_{\imcoord\in\imcoordset} \heatmap^k(\imcoord)\left(\imcoord - \mean^k\right)\left(\imcoord - \mean^k\right)^\trsp ~ \textrm{(SVD)},\\
\affine^k_\RtoX &= \begin{bmatrix}\U^k {\S^k}^\frac12, \mean^k\end{bmatrix}.
\end{align}
\noindent Here the singular value decomposition (SVD) approach to computing PCA \cite{wall2003singular} is used, Eq. (\ref{eq:pca}) decomposing the covariance of the heatmap into unitary matrices $\U^k$ and $\V^k$, and $\S^k$, the diagonal matrix of singular values.
We call this approach \emph{PCA-based}, in contrast to \emph{regression-based} for Eq.~(\ref{eq:shif}). Despite using the same region representation and encoder here, the encoded regions differ significantly (see Fig. \ref{fig:keypoints-qualitative}(a)), ours mapping to meaningful object parts such as the limbs of an articulated body, due to our novel foreground motion representation, described above. Note that in our \emph{PCA-based} approach, shear is not captured, therefore our transform is not fully affine, with only five degrees of freedom instead of six. Nevertheless, as we later show empirically, it captures sufficient motion, with shear being a less significant component of the affine transform for this task. The reference frame in both is used only as an intermediate coordinate frame between the source and driving image coordinate frames. However, here (in contrast to FOMM) it is not in fact abstract, corresponding to the coordinate frame where the heatmap is whitened (i.e. has zero mean and identity covariance); see Fig.~\ref{fig:framework}. $\affine_\DtoS^k$ is computed per Eq.~\eqref{eq:driving-to-source}.

\begin{table*}[t]
\centering
\resizebox{1\textwidth}{!}{
\begin{tabular}{c|ccc|ccc|ccc}
\toprule
  & \multicolumn{3}{c|}{5 regions} & \multicolumn{3}{c|}{10 regions} & \multicolumn{3}{c}{20 regions} \\
  & $\mathcal{L}_1$  & ({AKD}, {MKR}) & {AED} & $\mathcal{L}_1$  & ({AKD}, {MKR}) & {AED} & $\mathcal{L}_1$  & ({AKD}, {MKR}) & {AED}  \\  \toprule
    FOMM~\cite{fomm} & 0.062 & (7.34, 0.036) & 0.181 & 0.056 & (6.53, 0.033) & 0.172 & 0.062 & (8.29, 0.049) & 0.196 \\
    Ours w/o bg & 0.061 & (6.67, 0.030) & 0.175 &
         0.059 & ({\bf 5.55}, {\bf 0.026}) & 0.165 & 0.057 & (5.47, {\bf 0.026}) & 0.155
    \\
    Ours & \bf 0.049 & ({\bf 6.04}, {\bf 0.029}) & \bf 0.162 & \bf 0.047 & (5.59, 0.027) & \bf 0.152 & \bf 0.046 & 
    (\bf 5.17, \bf 0.026) & \bf 0.141 \\
    \bottomrule
\end{tabular}
}
\vspace{-0.2cm}
\caption{Comparing our model with FOMM~\cite{fomm} on TaiChiHD (256), for $\nregions = $ 5, 10 and 20. (Best result in bold.)}
\label{tab:pca_vs_regressed}
\vspace{-0.2cm}
\end{table*}

\newcommand{\bghomog}{\mathbf{\affine_\DtoS^0}}
\vspace{-0.1cm}
\subsection{Background motion estimation} 
\label{sec:bg}
\vspace{-0.1cm}
Background occupies a large portion of image. Hence even small background motion between frames, e.g.\ due to camera motion, negatively affects the animation quality. 
FOMM~\cite{fomm} does not treat background motion separately, therefore must model it using keypoints. This has two negative consequences: (i) additional network capacity is required, since several keypoints are used to model the background instead of the foreground; (ii) overfitting to the training set, since these keypoints focus on specific parts of the background, which may not appear in the test set. Hence, we additionally predict an affine background transformation, $\bghomog$, using an encoder network assuming $\sourceframe$ and $\drivingframe$ as input and predicting six real values, $a_1,..,a_6$, such that
$\bghomog = \left[\begin{smallmatrix} a_1, a_2, a_3 \\ a_4, a_5, a_6\end{smallmatrix}\right]$.
Since our framework is unsupervised, the background network can include parts of the foreground into the background motion. In practice this does not happen, since it is easier for the network to use a more appropriate \emph{PCA-based} motion representation for the foreground. It is also simpler for the network to use $\sourceframe$ and $\drivingframe$ to predict background movement, instead of encoding it in the heatmaps modelling the foreground. We verify this empirically, demonstrating that the proposed motion representations can separate background and foreground in a completely unsupervised manner (see Fig.~\ref{fig:copart} and Sec.~\ref{sec:co-part} for comparisons).
\vspace{-0.1cm}
\subsection{Image generation}
\vspace{-0.1cm}
Similarly to FOMM~\cite{fomm}, we render the target image in two stages: a pixel-wise flow generator converts coarse motions to dense optical flow, then the encoded features of the source are warped according to the flow, followed by inpainting the missing regions. The input of the dense flow predictor is a $\height\times\width\times(4\nregions+3)$ tensor, with four channels per region, three for the source image warped according to the region's affine transformation, and one for a heatmap of the region, which is a gaussian approximation of $\heatmap^k$, and a further three channels for the source image warped according to the background's affine transformation. In contrast to FOMM, which uses constant variances, we estimate covariances from heatmaps. Our dense optical flow is given by (cf.\ Eq.~(\ref{eq:flow})):
\begin{equation}
    \opticalflow(z) = \sum_{k=0}^K \assignment^k(z) \affine^k_\DtoS \begin{bmatrix} z \\ 1\end{bmatrix}.
\end{equation}
The predicted optical flow and confidence map are used as per FOMM~\cite{fomm}. However in contrast to FOMM~\cite{fomm}, but similar to Monkey-Net~\cite{siarohin2018animating}, here deformable skip connections~\cite{siarohin2018deformable} are used between the encoder and decoder.
\vspace{-0.1cm}
\subsection{Training}
\label{sec:training}
\vspace{-0.1cm}
The proposed model is trained end-to-end using a reconstruction loss in the feature space of the pretrained VGG-19 network~\cite{johnson2016perceptual, wang2017high}. We use a multi-resolution reconstruction loss from previous work~\cite{fomm, tang2018dual}:
\begin{equation}
    \mathcal{L}_{\mathrm{rec}}(\mathbf{\hat{D}}, \mathbf{D})  = \sum_{l} \sum_{i} \left|\mathrm{V}_i(\mathrm{F}_l \odot \mathbf{\hat{D}}) - \mathrm{V}_i(\mathrm{F}_l \odot \mathbf{D}) \right|,
\end{equation}
where $\mathbf{\hat{D}}$ is the generated image, $\mathrm{V}_i$ is the $i^{\textrm{th}}$-layer of the VGG-19 pretrained network, and $\mathrm{F}_l$ is a downsampling operator. Per FOMM~\cite{fomm}, we also use an equivariance loss,
\begin{equation}
    \mathcal{L}_\mathrm{eq} = \left|\affine^k_\RtoX - \tilde{A} \affine^k_{\tilde{\imX} \leftarrow \refframe}\right|,
\end{equation}
where $\tilde{\imX}$ is image $\imX$ transformed by $\tilde{A}$, and $\tilde{A}$ is some random geometric transformation.
The final loss is the sum of terms, $\mathcal{L}=\mathcal{L}_\mathrm{rec} + \mathcal{L}_\mathrm{eq}$.

\vspace{-0.1cm}
\subsection{Animation via disentanglement}
\label{sec:animation}
\vspace{-0.1cm}
Image animation using both \emph{standard} and \emph{relative} methods has limitations. The \emph{standard} method directly transfers object shape from the driving frame into the generated video, 
while \emph{relative} animation is only applicable to a limited set of inputs, e.g.\ it requires that objects be in the same pose in the source $\sourceframe$ and initial driving $\drivingframe_1$ frames. To address this, we learn disentangled shape and pose encoders, as shown in Fig.~\ref{fig:keypoints-qualitative}(b). The pose encoder takes in the set of \emph{driving} motions, $\{\affine^k_\RtoD\}_{k=1}^K$, while the shape encoder takes in the set of \emph{source} motions, $\{\affine^k_\RtoS\}_{k=1}^K$. A decoder then uses the concatenated latent representations (each in $\real^{64}$) of these two encoders, to produce a set of modified driving motions, $\{\tilde{\affine}^k_\RtoD\}_{k=1}^K$ encoding the motion of the former and the shape of the latter. These are then used to render the output.


The encoders and the decoder are implemented using fully connected layers, and trained separately from (and after) other blocks, using an $\mathcal{L}_1$ reconstruction loss on the motion parameters. As with earlier training, source and driving frames come from the same video (i.e.\ the object has the same shape), therefore to ensure that shape comes from the correct branch, random horizontal and vertical scaling deformations are applied to the driving motions during training, as shown in Fig.~\ref{fig:keypoints-qualitative}(b). This forces shape information to come from the other (shape) branch. However, since the shape branch has a different pose, the pose must still come from the pose branch. Thus shape and pose are disentangled. The deformations are not applied at test time.


\newcommand{\features}{f}
\newcommand{\nchannels}{N}
\newcommand{\nimages}{I}
\vspace{-0.2cm}
\section{Evaluation}
\vspace{-0.2cm}
We now discuss the datasets, metrics and experiments used to evaluate the proposed method. Later we compare with prior work, as well as ablate our contributions.

\begin{figure*}[t]
    \centering
    \begin{overpic}[width=\linewidth,trim=0 0 0 0,clip]{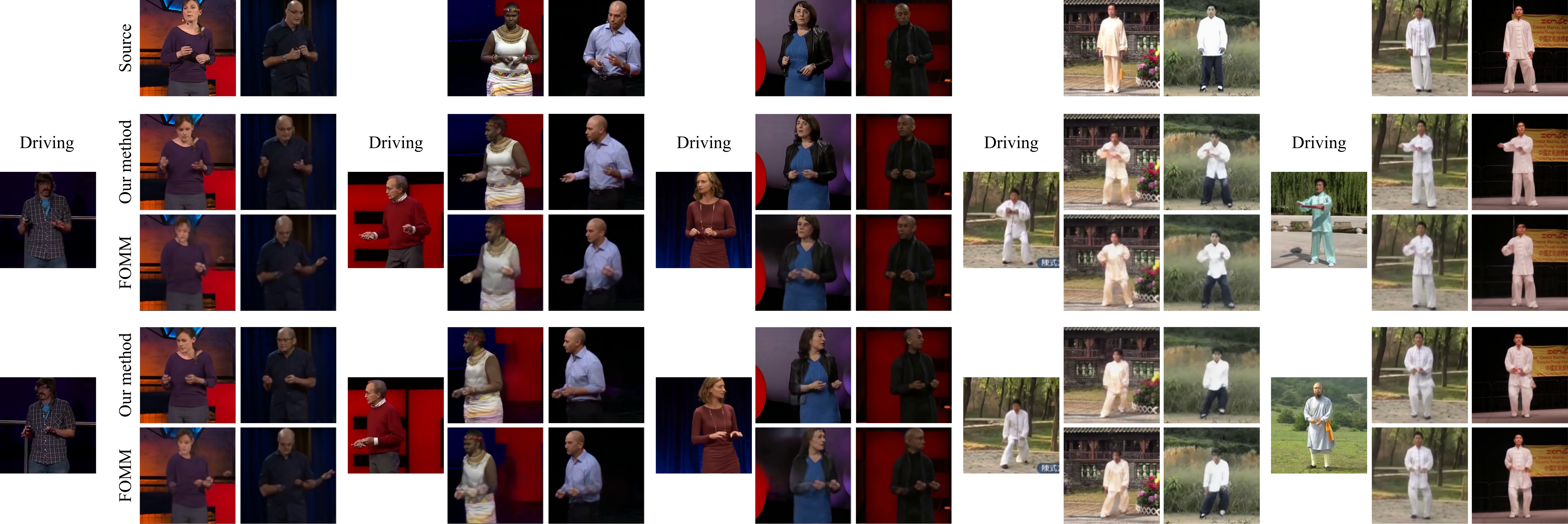}
    \end{overpic}
    \vspace{-0.65cm}
    \caption{\textbf{Qualitative comparisons.} We show representative examples of articulated animation using our method and FOMM~\cite{fomm}, on two datasets of articulated objects: TED-talks (left) and TaiChiHD (right). Zoom in for greater detail.}
    \label{fig:qualitative}
\end{figure*}

\subsection{Toy Motion Representation Experiment} 
\label{sec:toy}
\begin{figure}[h]
    \centering
    \includegraphics[width=0.9\linewidth]{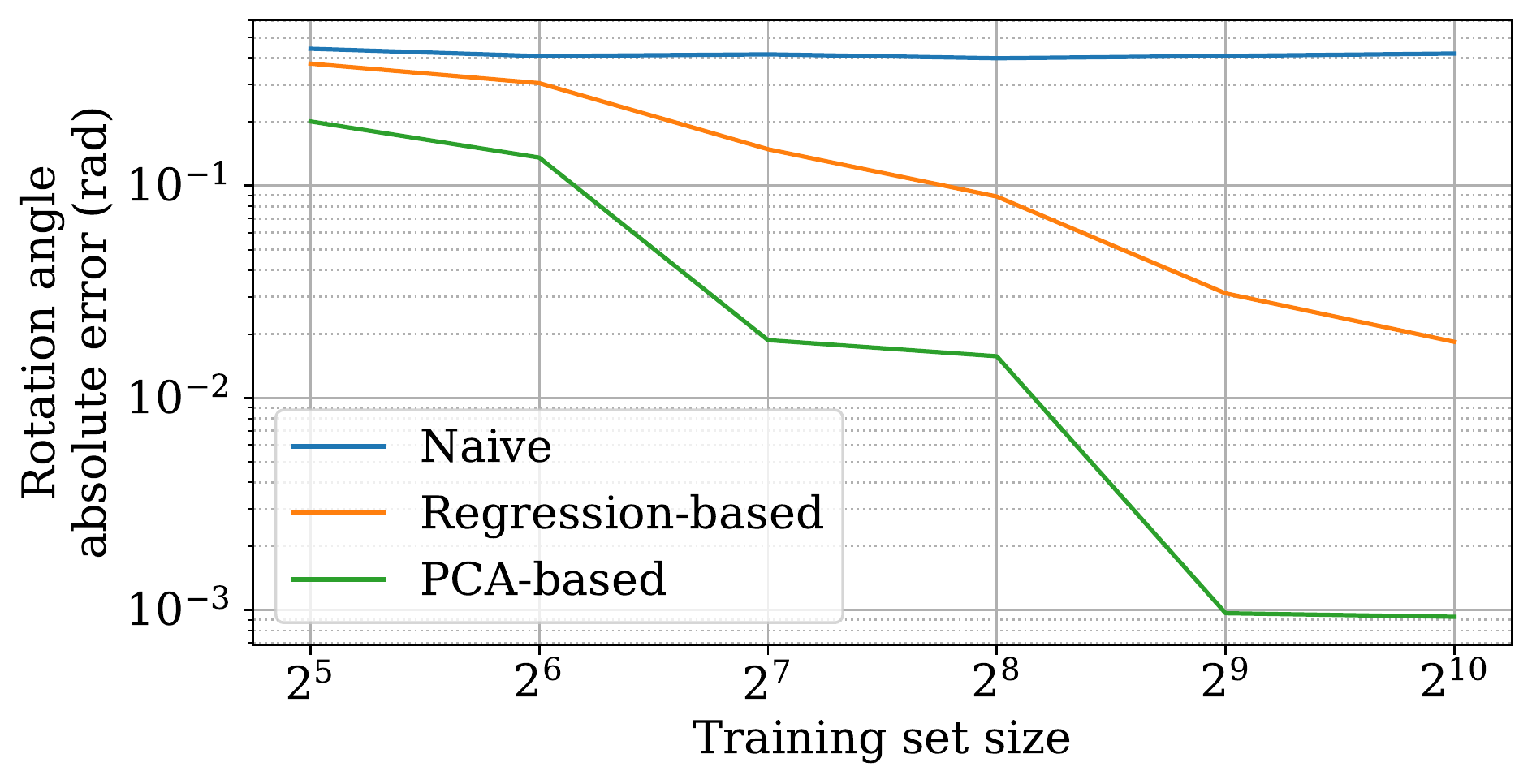}
    \vspace{-0.4cm}
    \caption{Mean test-time absolute rotation error, as a function of training set size.
    }
    \label{fig:rectangles-result}
    \vspace{-0.6cm}
\end{figure}

To demonstrate the benefit of the proposed PCA-based motion representation, we devise an experiment on rotated rectangles (see Sup. Mat.): the task is to predict the rotation angle of a rectangle in an image. To fully isolate our contribution, we consider a supervised task, where three different architectures learn to predict angles under the $\mathcal{L}_1$ loss. The first, a Naive architecture, directly regresses the angle using an encoder-like architecture. The second is Regression-based, as in FOMM~\cite{fomm}. The third uses our PCA-based approach (see Sup. Mat.). Test results are presented in Fig.~\ref{fig:rectangles-result}, against training set size. The Naive baseline struggles to produce meaningful results for any size of training set, while Regression-based performance improves with more data. However, the PCA-based approach significantly improves accuracy over the Regression-based one, being over an order of magnitude better with a large number of samples. This shows that it is significantly easier for the network to infer geometric parameters of the image, such as angle, using our proposed PCA-based representation.

\subsection{Benchmarks}
We evaluate our method on several benchmark datasets for animating human faces, bodies and animated cartoons. 
Each dataset has separate training and test videos. The datasets are as follows:
\begin{itemize*}
\item\emph{VoxCeleb}~\cite{Nagrani17} consists of interview videos of different celebrities. We extract square, face regions and downscale them to $256 \times 256$, following FOMM~\cite{fomm}. The number of frames per video ranges from 64 to 1024.
\item\emph{TaiChiHD}~\cite{fomm} consists of cropped videos of full human bodies performing Tai Chi actions. We evaluate on two resolutions of the dataset: $256 \times 256$ (from FOMM~\cite{fomm}), and a new, $512 \times 512$ subset, removing videos lacking sufficient resolution to support that size.
\item\emph{MGif}~\cite{siarohin2018animating} is a dataset of \emph{.gif} files, that depicts 2D cartoon animals. The dataset was collected using google searches.
\item\emph{TED-talks} is a new dataset, collected for this paper in order to demonstrate the generalization properties of our model. We cropped the upper part of the human body from the videos, downscaling to $384 \times 384$. The number of frames per video ranges from 64 to 1024.
\end{itemize*}

Since video animation is a relatively new problem, there are not currently many effective ways of evaluating it. For quantitative metrics, prior works~\cite{wiles2018x2face,siarohin2018animating,fomm} use video reconstruction accuracy as a proxy for image animation quality. We adopt the same metrics here:
\begin{itemize*}
    \item $\mathcal{L}_1$ error is the mean absolute difference between reconstructed and ground-truth video pixel values. 
    \item \emph{Average keypoint distance} (AKD) and \emph{missing keypoint rate} (MKR) evaluate the difference between poses of reconstructed and ground truth video. Landmarks are extracted from both videos using public, body~\cite{cao2017realtime} (for TaiChiHD and TED-talks) and face~\cite{Bulat_2017_ICCV} (for VoxCeleb) detectors. AKD is then the average distance between corresponding landmarks, while MKR is the proportion of landmarks present in the ground-truth that are missing in the reconstructed video.
    \item \emph{Average Euclidean distance} (AED) evaluates how well identity is preserved in reconstructed video. Public re-identification networks for bodies~\cite{hermans2017defense} (for TaiChiHD and TED-talks) and faces~\cite{amos2016openface} extract identity from reconstructed and ground truth frame pairs, then we compute the mean $\mathcal{L}_2$ norm of their difference across all pairs.
\end{itemize*}


\begin{table*}[t]
    \centering
    \resizebox{1\textwidth}{!}{
    \begin{tabular}{c|ccc|ccc|ccc|ccc|c}
    \toprule
      & \multicolumn{3}{c|}{TaiChiHD (256)} & \multicolumn{3}{c|}{TaiChiHD (512)} & \multicolumn{3}{c|}{TED-talks} & \multicolumn{3}{c|}{VoxCeleb} & MGif \\
      & $\mathcal{L}_1$  & ({AKD}, {MKR}) & {AED} & $\mathcal{L}_1$  & ({AKD}, {MKR}) & {AED} & $\mathcal{L}_1$  & ({AKD}, {MKR}) & {AED} & $\mathcal{L}_1$  & {AKD} & {AED}  & $\mathcal{L}_1$ \\  \toprule
        X2Face & 0.080 & (17.65, 0.109) & 0.27 &-&-&-&-&-&-&
        0.078 & 7.69 & 0.405 &-\\
        Monkey-Net & 0.077 & (10.80, 0.059) & 0.228 &-&-&-&-&-&-&   
        0.049 &  1.89 & 0.199 &-\\ 
        FOMM & 0.056 & (6.53, 0.033) & 0.172 & 0.075 & (17.12, 0.066)  & 0.203 & 0.033 & (7.07, 0.014) & 0.163 & 0.041 & \bf 1.27 & 0.134 & 0.0223 \\
        Ours & \bf 0.047 &	({\bf 5.58}, {\bf 0.027}) &	\bf 0.152 & \bf 0.064 & ({\bf 13.86}, {\bf 0.043}) & \bf 0.172 & \bf 0.026 & ({\bf 3.75}, {\bf 0.007}) & {\bf 0.114} & \bf 0.040 & 1.28 & \bf 0.133 & 
        \bf  0.0206\\
        \bottomrule
    \end{tabular}}
    \vspace{-0.2cm}
    \caption{Video reconstruction: comparison with the state of the art on five different datasets. For all methods we use $\nregions=10$ regions. (Best result in bold.)}
    \label{tab:sota}
    \vspace{-0.4cm}
\end{table*}

\subsection{Comparison with the state of the art}
We compare our method with the current state of the art for unsupervised animation, FOMM~\cite{fomm}, on both reconstruction (the training task) and animation (the test-time task). We used an extended training schedule compared to \cite{fomm}, with 50\% more iterations. To compare fairly with FOMM~\cite{fomm}, we also re-trained it with the same training schedule. Also, for reference, we include comparisons with X2Face~\cite{wiles2018x2face} and Monkey-Net~\cite{siarohin2018animating} on video reconstruction.

\paragraph{Reconstruction quality}
Quantitative reconstruction results are reported in Table~\ref{tab:sota}. We first show that our method reaches state-of-the-art results on a dataset with non-articulated objects such as faces. Indeed, when compared with FOMM~\cite{fomm} on VoxCeleb, our method shows on-par results. The situation changes, however, when articulated objects are considered, such as human bodies in TaiChiHD and TED-talks datasets, on which our improved motion representations boost all the metrics. The advantage over the state of the art holds at different resolutions, for TaiChiHD (256), TaiChiHD (512) and TED-talks, as well as for different numbers of selected regions (discussed later). 

\paragraph{Animation quality}
\vspace{-0.2cm}

Fig.~\ref{fig:keypoints-qualitative}(a,c) \& \ref{fig:qualitative} show selected and representative animations respectively, using our method and FOMM~\cite{fomm}, on articulated bodies. Here for FOMM~\cite{fomm} we use the \emph{standard} method, while ours uses animation via disentanglement. The results show clear improvements, in most cases, in animation quality, especially of limbs.

Animation quality was evaluated quantitatively through a user preference study similar to that of \cite{fomm}. AMT users were presented with the source image, driving video, and the output from our method and FOMM~\cite{fomm}, and asked which of the two videos they preferred. 50 such videos were evaluated, by 50 users each, for a total of 2500 preferences per study.
The results further support the reconstruction scores in Tab.~\ref{tab:sota}. When the animated object is not articulated (VoxCeleb), the method delivers results comparable to the previous work, e.g.\ 52\% preference in favour of our method. However, when bodies are animated (TaiChiHD \& TED-talks), FOMM~\cite{fomm} fails to correctly detect and animate the articulated body parts such as hands. Our method renders them in the driving pose even for extreme cases, leading to a high preference in favor of it (see Tab.~\ref{tab:user}, middle column). Moreover since it is not possible to demonstrate the benefits of the animation via disentanglement using reconstruction metrics, we run an additional user study to compare our method with \emph{standard} animation and animation via disentanglement. Since animation via disentanglement preserves shape of the object much better (see Fig.~\ref{fig:keypoints-qualitative}(c)), users prefer it more often. It especially pronoun in case of the TED-talks dataset, since shape of the objects differs more in that case (see Tab.~\ref{tab:user}, last column). 

\begin{table}[t]
\centering
    \resizebox{1\linewidth}{!}{
    \begin{tabular}{c|c|c}
        \toprule
        Dataset &  Our vs FOMM (\%) & Our vs Standard (\%) \\
        \midrule
        TaiChiHD (256) & 83.7\% & 53.6\% \\
        TED-talks & 96.6\% & 65.6\%\\ 
        \bottomrule
        
        \end{tabular}
    }
    \caption{User study: second column - the proportion (\%) of users that prefer our method over FOMM~\cite{fomm}; third column - the proportion (\%) of users that prefer animation via disentanglement over \emph{standard} animation for our model.}
    \label{tab:user}
    \vspace{-0.6cm}
\end{table}

Finally, we applied animation from a TED-talks video to a photograph of George Washington, shown in Fig.~\ref{fig:teaser}, demonstrating animation of out-of-domain data. 

\subsection{Ablations}

\begin{table}[t]
\centering
    \begin{tabular}{c|ccc}
        \toprule
        & $\mathcal{L}_1$ & ({AKD}, {MKR}) & {AED} \\  \midrule
        No pca or bg model & 0.060 & (6.14, 0.033) & 0.163 \\
        No pca & 0.056 & (9.58, 0.034) & 0.206 \\
        No bg model & 0.059 & ({\bf 5.55}, {\bf 0.026}) & 0.165 \\
        Full method & \bf 0.048 & (5.59, 0.027) & \bf 0.152\\
        \bottomrule
    \end{tabular}
    \caption{Ablation study on TaiChiHD (256) dataset with $\nregions=10$. (Best result in bold.)}
    \label{tab:ablation}
    \vspace{-0.2cm}
\end{table}

\begin{figure}[t]
    \centering
    \includegraphics[width=0.9\linewidth]{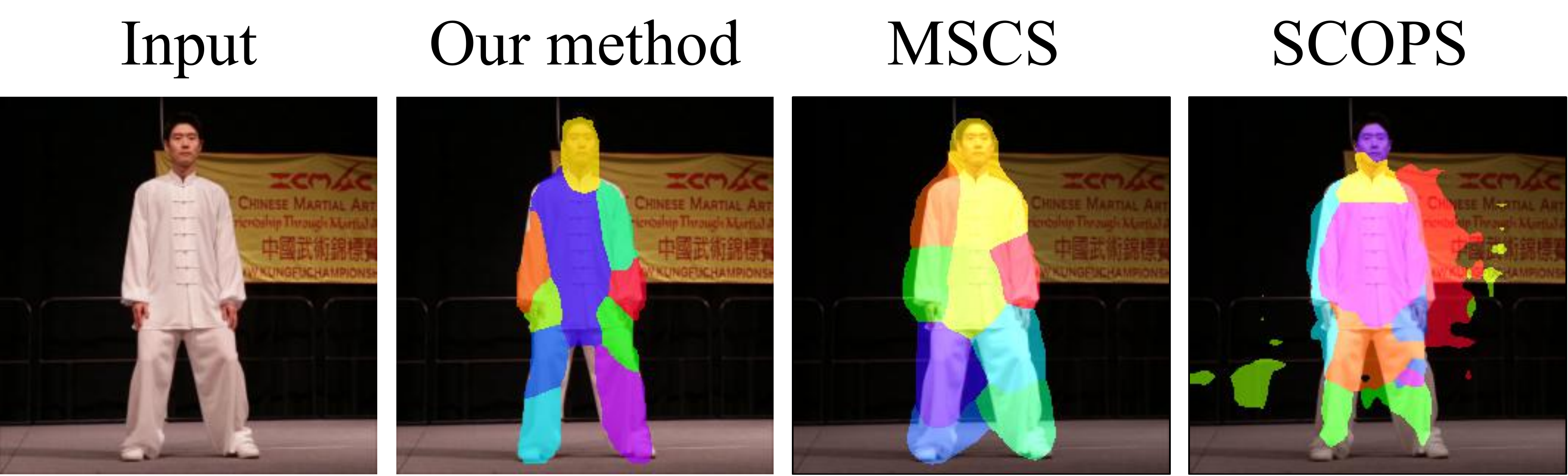}
    \caption{Qualitative co-part segmentation comparison with recent methods.}
    \label{fig:copart}
    \vspace{-0.4cm}
\end{figure}

In order to understand how much benefit each of our contributions bring, we ran a number of ablation experiments, detailed in Tab.~\ref{tab:ablation}.
\paragraph{PCA-based vs.~regression-based representations} First we compare the PCA-based motion model with the previous, regression-based one~\cite{fomm}. From the qualitative heatmap depictions in Fig.~\ref{fig:keypoints-qualitative}(a), we observe that the regression-based method localizes one edge of each corresponding part, while our method predicts regions that roughly correspond to the segmentation of the object into its constituent, articulated parts. This meaningful segmentation arises completely unsupervised.

From Tab.~\ref{tab:pca_vs_regressed} we note that adding the PCA-based representation alone (second row) had marginal impact on the $\mathcal{L}_1$ score (dominated by the much larger background region), but it had a much larger impact on other metrics, which are more sensitive to object-part-related errors on articulated objects. This is corroborated by Tab.~\ref{tab:ablation}. Moreover, we observed that when our PCA-based formulation is not used, the network encodes some of the movement into the background branch, leading to significant degradation of keypoint quality, which in turn leads to degradation of the AKD and AED scores (No-pca, Tab.~\ref{tab:ablation}).

We intuit that PCA-based estimation both captures regions and improves performance because it is much easier for the convolutional network to assign pixels of an object part to the corresponding heatmap than to directly regress motion parameters to an abstract reference frame. This is borne out by our toy experiment (sec.\ \ref{sec:toy}). In order to estimate the heatmap, it need only learn all appearances of the corresponding object part, while regression-based networks must learn the joint space of all appearances of a part in all possible geometric configurations (e.g.~rotated, scaled etc.).

One of the most important hyper-parameters of our model is the number of regions, $\nregions$. The qualitative and quantitative ablations of this parameter are shown in Fig.~\ref{fig:keypoints-qualitative}(a) and Tab.~\ref{tab:pca_vs_regressed} respectively. We can observe that, while the regression-based representation fails when the number of keypoints grows to 20, our PCA-based representation scales well with the number of regions.

\paragraph{Modeling background motion} Background motion modeling significantly lowers $\mathcal{L}_1$ error (see Tab.~\ref{tab:ablation}, Full method vs.\ No bg Model). Since background constitutes a large portion of the image, and $\mathcal{L}_1$ treats all pixels equally, this is to be expected. AED was also impacted, suggesting that the identity representation captures some background appearance. Indeed, we observe (Fig.~\ref{fig:keypoints-qualitative}(a), \emph{second row}) that having no background model causes a reduction in region segmentation quality. However, since AKD \& MKR metrics evaluate object pose only, they are not improved by background modelling.

\subsection{Co-part Segmentation}
\label{sec:co-part}
While designed for articulated animation, our method produces meaningful object parts. To evaluate this capability of our method, we compare it against two recent unsupervised co-part segmentation works: MSCS~\cite{siarohin2020motionsupervised} and SCOPS~\cite{Hung_2019_CVPR}. Following MSCS, we compute \emph{foreground segmentation IoU} scores on TaiChiHD. Despite not being optimized for this task, our method achieves superior performance reaching 0.81 \emph{IoU} vs.\ 0.77 for MSCS~\cite{siarohin2020motionsupervised} and 0.55 for SCOPS~\cite{Hung_2019_CVPR}. See Fig.~\ref{fig:copart} for qualitative results.

\vspace{-0.2cm}
\section{Conclusion}
\vspace{-0.2cm}
We have argued that previous unsupervised animation frameworks' poor results on articulated objects are due to their representations. We propose a new, PCA-based, region motion representation, that we show both makes it easier for the network to learn region motion, and encourages it to learn semantically meaningful object parts. In addition, we propose a background motion estimation module to decouple foreground and background motion. Qualitative and quantitative results across a range of datasets and tasks demonstrate several key benefits: improved region distribution and stability, improved reconstruction accuracy and user perceived quality, and an ability to scale to more regions. We also introduce a new, more challenging dataset, TED-talks, for benchmarking future improvements on this task.

While we 
show some results on out of domain data (Fig.~\ref{fig:teaser}), generalization remains a significant challenge to making this method broadly practical in articulated animation of inanimate objects.

\clearpage

\renewcommand{\thesection}{\Alph{section}}
\setcounter{section}{0}

In this supplementary material we report additional details on the toy experiment in Sec.~\ref{sec:motivation}.  In Sec.~\ref{sec:copart} we provide additional details for the co-part segmentation experiment. We provide additional implementation details in Sec.~\ref{sec:impl}.
Additionally in Sec.~\ref{sec:bg_movement} we visually demonstrate the ability of the model to control the background. Finally in Sec.~\ref{sec:ted} we describe the TED-talks data collection procedure. 

\section{Toy Experiment Details}
\label{sec:motivation}
\begin{figure}[h]
    \centering
    \includegraphics[width=\linewidth]{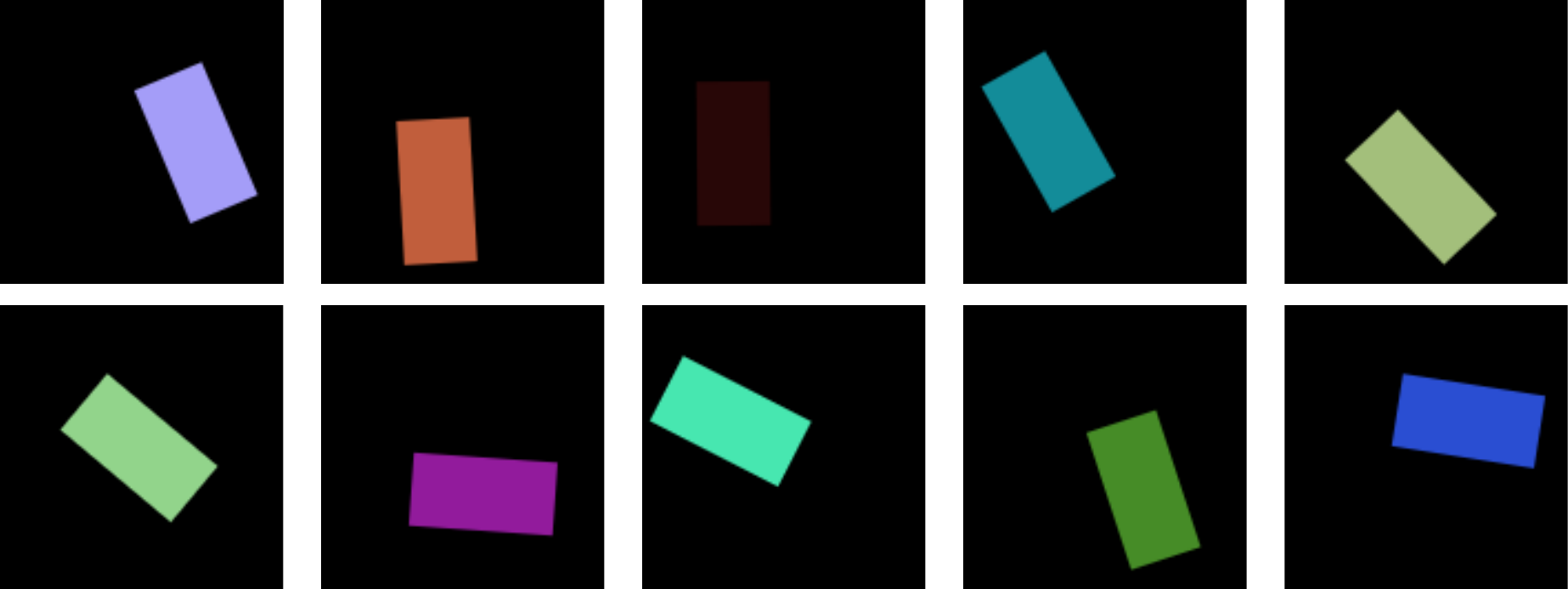}
    \caption{Examples of synthetic rectangle dataset.}
    \label{fig:rectangles}
\end{figure}

The rotated rectangles dataset consists of images of rectangles randomly rotated from $0^{\circ}$ to $90^{\circ}$, along with labels that indicate the angle of rotation. The rectangles have different, random colors. Visual samples are shown in Fig.~\ref{fig:rectangles}.

We tested three different networks: Naive, Regression-based and PCA-based. The Naive network directly predicts an angle from an image using an encoder and a fully-connected layer. Regression-based is similar to FOMM~\cite{fomm}; the angle is regressed per pixel an using hourglass network, and pooled according to heatmap weights predicted using the same hourglass network. PCA-based is our method described in Sec.~\ref{sec:regions} we predict the heatmap using an hourglass network, PCA is performed according to Eq.~\eqref{eq:pca}, and the angle is computed from matrix $U$ as $\arctan(U_{10}/U_{00})$.

Each of the networks was trained, on subsets of the dataset of varying sizes, to minimize the $\mathcal{L}_1$ loss between predicted and ground truth rotation angle. All models were trained for $100$ epochs, with batch size $8$. We used the Adam optimizer, with a learning rate of $10^{-4}$. We varied the size of the training set from $32$ to $1024$. Results, on a separate, fixed test set of size $128$, were then computed, shown in Fig.~\ref{fig:rectangles-result}.

\section{Co-part segmentation details}
\label{sec:copart}

To perform co-part segmentation we use $\heatmap^k$. 
A pixel $z$ is assigned to a part that has the maximum response of  $\heatmap^k$ in that pixel, e.g $\mathrm{argmax}_k\heatmap^k(z)$
Moreover since region predictor did not explicitly predict background region, we assign pixel $z$ to the background iff $\sum_k  \heatmap^k(z) < 0.001$. 
We demonstrate additional qualitative comparisons with MSCS~\cite{siarohin2020motionsupervised} and SCOPS~\cite{Hung_2019_CVPR} in Fig.~\ref{fig:copar-additional}. Fig.~\ref{fig:copar-additional} shows that our method produces more meaningful co-part segmentations compared to SCOPS~\cite{Hung_2019_CVPR}. Furthermore, our method separates the foreground object from the background more accurately than MSCS~\cite{siarohin2020motionsupervised}.

\begin{figure}[h]
    \centering
    \includegraphics[width=\linewidth]{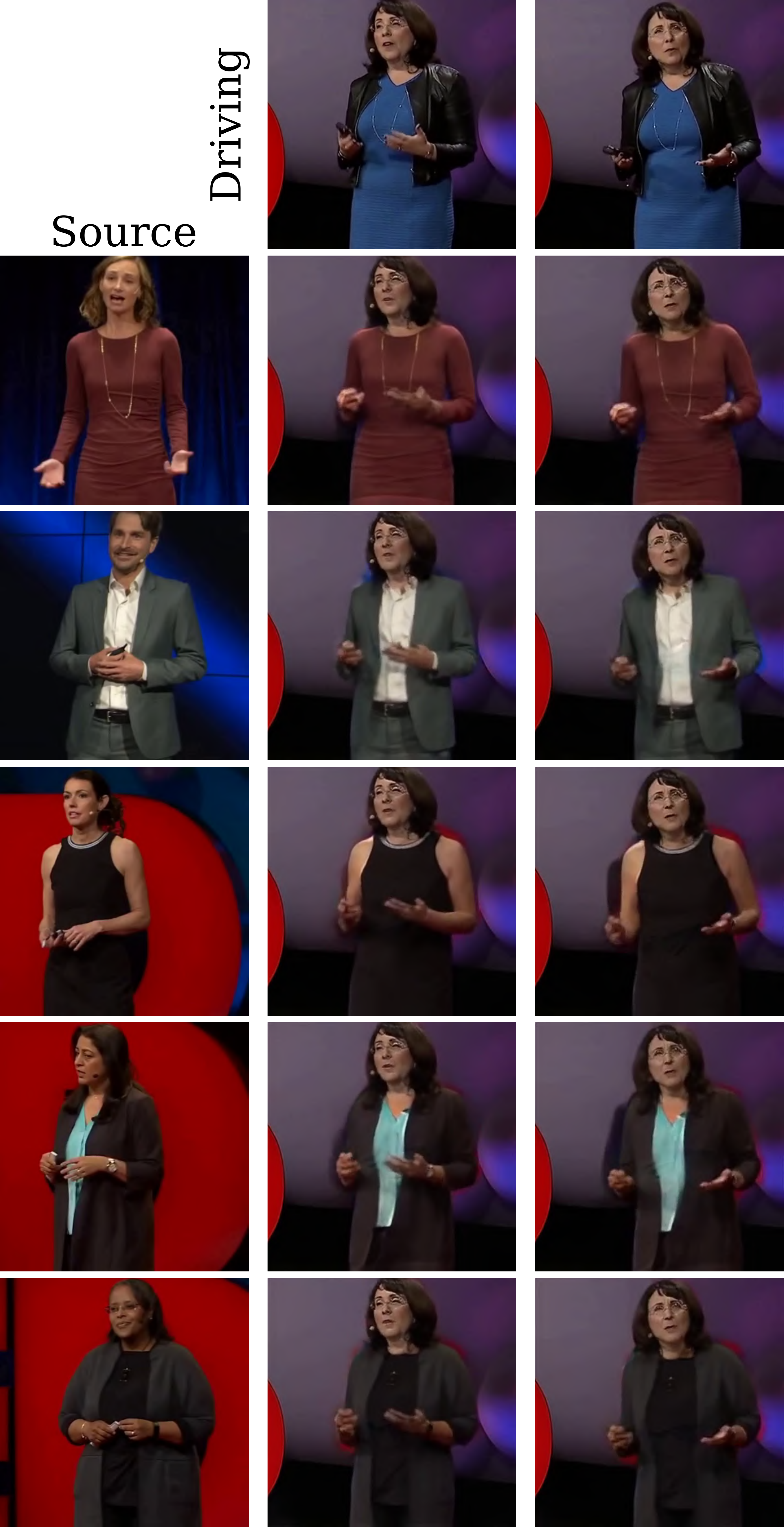}
    \caption{Examples of cloth swap performed using our model. First column depicts sources from which cloth is taken, while the first row shows a driving video to which we put the cloth. Rest demonstrates images generated with our model.}
    \label{fig:tryon}
\end{figure}
Similarly to MSCS~\cite{siarohin2020motionsupervised} we can exploit produced segmentations in order to perform a part swap. In Fig.~\ref{fig:tryon} we copy the cloth from the person in the source image on to the person in the driving video. 

\begin{table*}[h]
    \centering
    \def\arraystretch{0.9}
    \resizebox{\linewidth}{!}{
    \begin{tabular}{cccc}
        Input & Ours & SCOPS~\cite{Hung_2019_CVPR} & MSCS~\cite{siarohin2020motionsupervised}\\\\
        \includegraphics[width=0.23\columnwidth]{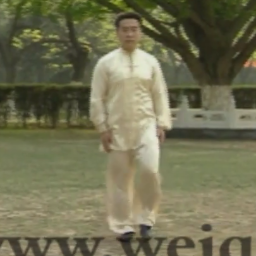} & \includegraphics[width=0.23\columnwidth]{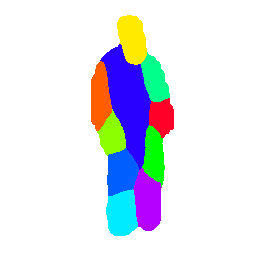}\includegraphics[width=0.23\columnwidth]{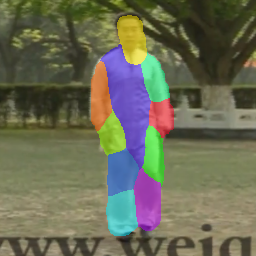} & \includegraphics[width=0.23\columnwidth]{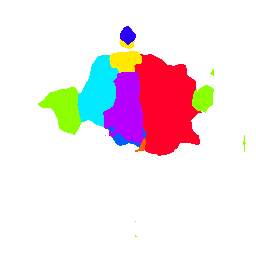}\includegraphics[width=0.23\columnwidth]{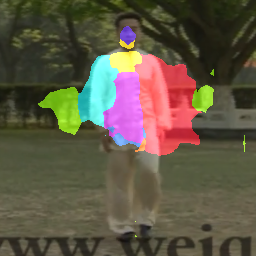} & \includegraphics[width=0.23\columnwidth]{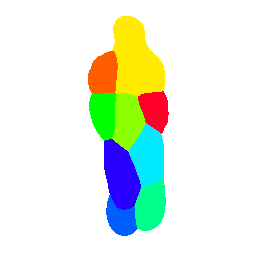}\includegraphics[width=0.23\columnwidth]{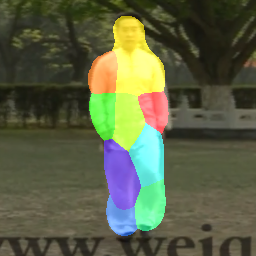}\\
        \includegraphics[width=0.23\columnwidth]{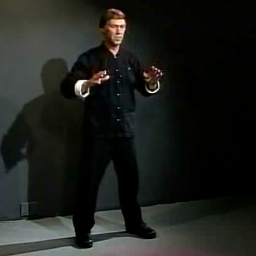} & \includegraphics[width=0.23\columnwidth]{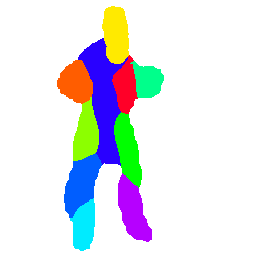}\includegraphics[width=0.23\columnwidth]{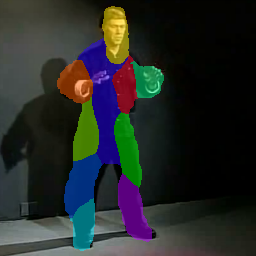} & \includegraphics[width=0.23\columnwidth]{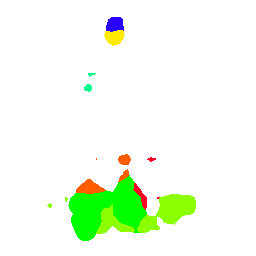}\includegraphics[width=0.23\columnwidth]{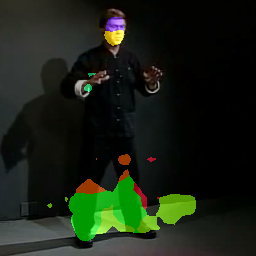} & \includegraphics[width=0.23\columnwidth]{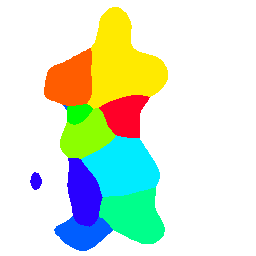}\includegraphics[width=0.23\columnwidth]{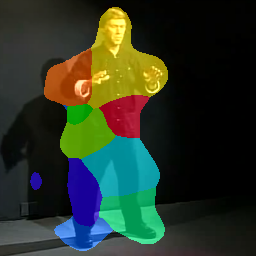}\\
        
        \includegraphics[width=0.23\columnwidth]{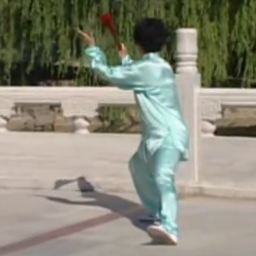} & \includegraphics[width=0.23\columnwidth]{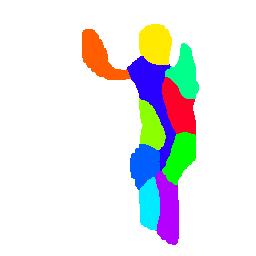}\includegraphics[width=0.23\columnwidth]{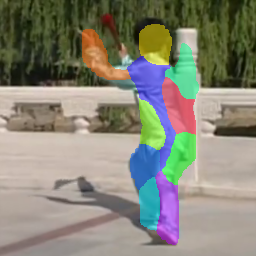} & \includegraphics[width=0.23\columnwidth]{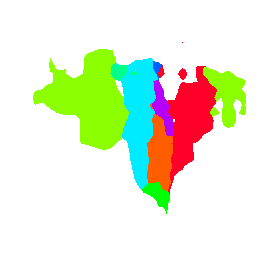}\includegraphics[width=0.23\columnwidth]{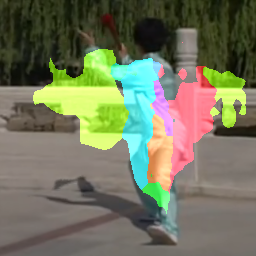} & \includegraphics[width=0.23\columnwidth]{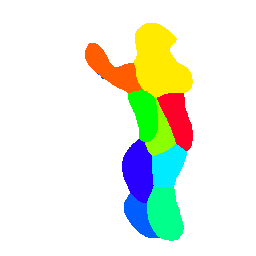}\includegraphics[width=0.23\columnwidth]{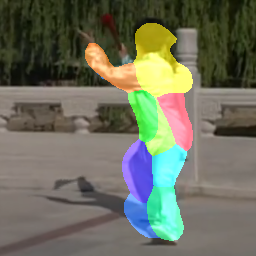}\\
        
       \includegraphics[width=0.23\columnwidth]{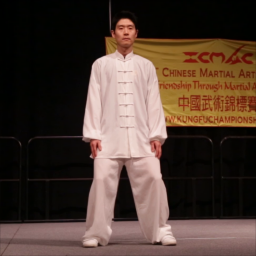} & \includegraphics[width=0.23\columnwidth]{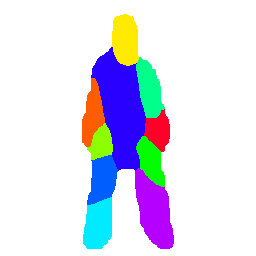}\includegraphics[width=0.23\columnwidth]{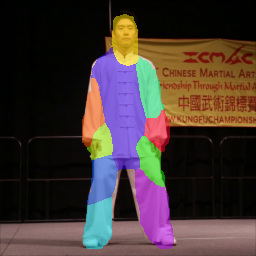} & \includegraphics[width=0.23\columnwidth]{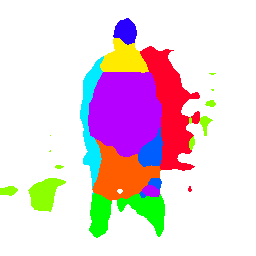}\includegraphics[width=0.23\columnwidth]{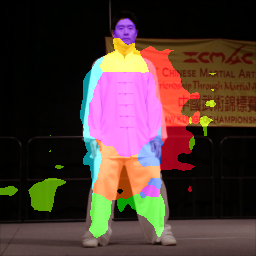} & \includegraphics[width=0.23\columnwidth]{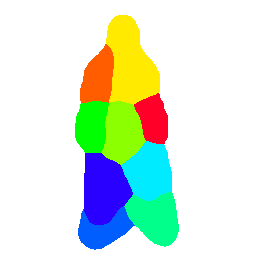}\includegraphics[width=0.23\columnwidth]{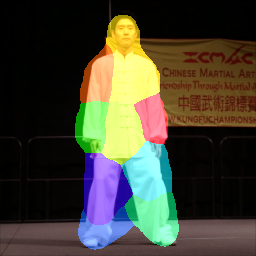}\\
        
        \includegraphics[width=0.23\columnwidth]{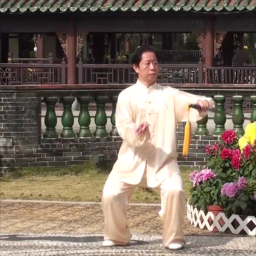} & \includegraphics[width=0.23\columnwidth]{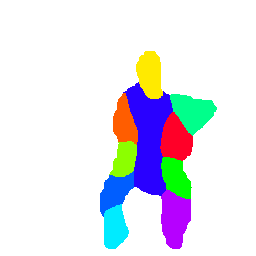}\includegraphics[width=0.23\columnwidth]{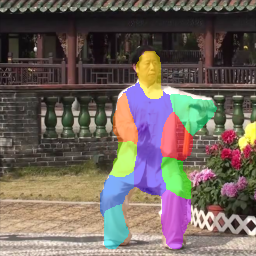} & \includegraphics[width=0.23\columnwidth]{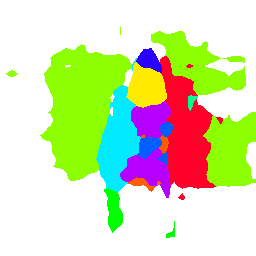}\includegraphics[width=0.23\columnwidth]{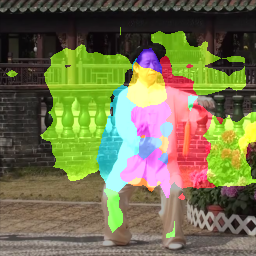} & \includegraphics[width=0.23\columnwidth]{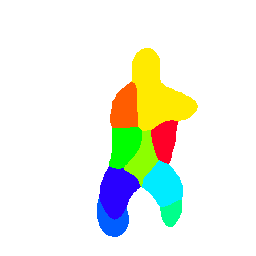}\includegraphics[width=0.23\columnwidth]{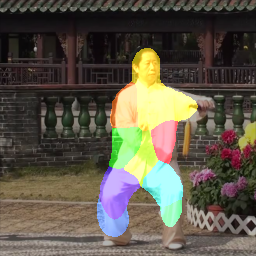}\\
        
        \includegraphics[width=0.23\columnwidth]{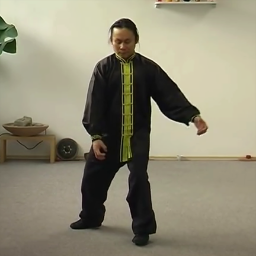} & \includegraphics[width=0.23\columnwidth]{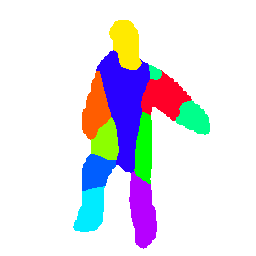}\includegraphics[width=0.23\columnwidth]{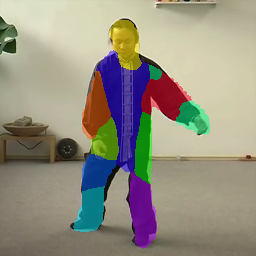} & \includegraphics[width=0.23\columnwidth]{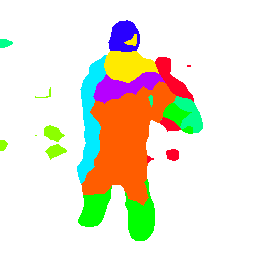}\includegraphics[width=0.23\columnwidth]{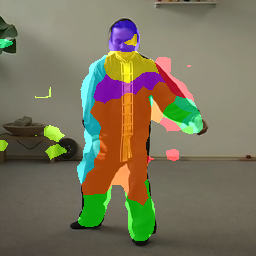} & \includegraphics[width=0.23\columnwidth]{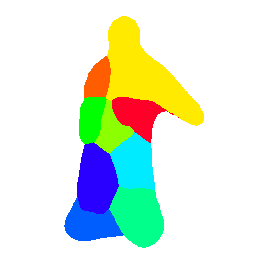}\includegraphics[width=0.23\columnwidth]{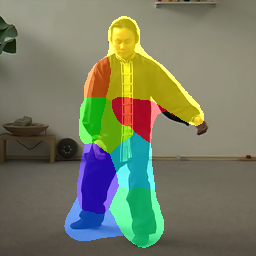}\\
        
        \includegraphics[width=0.23\columnwidth]{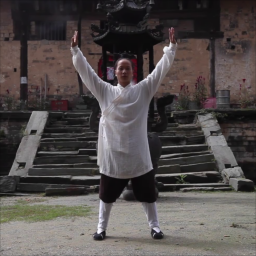} & \includegraphics[width=0.23\columnwidth]{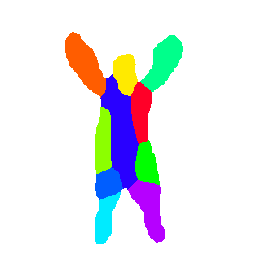}\includegraphics[width=0.23\columnwidth]{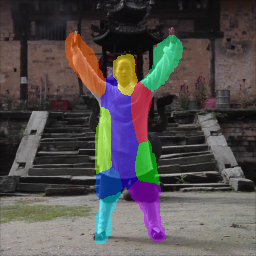} & \includegraphics[width=0.23\columnwidth]{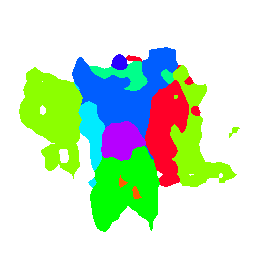}\includegraphics[width=0.23\columnwidth]{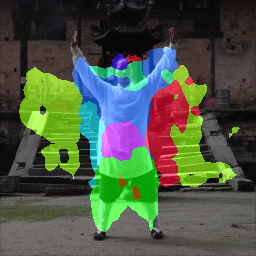} & \includegraphics[width=0.23\columnwidth]{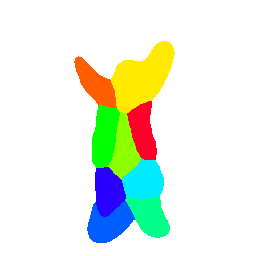}\includegraphics[width=0.23\columnwidth]{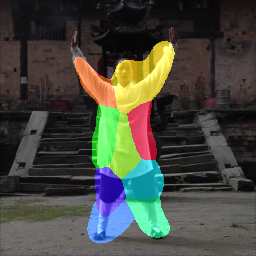}\\
        
        \includegraphics[width=0.23\columnwidth]{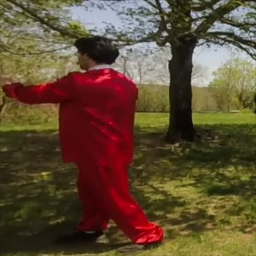} & \includegraphics[width=0.23\columnwidth]{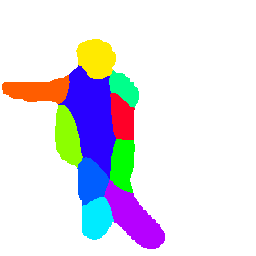}\includegraphics[width=0.23\columnwidth]{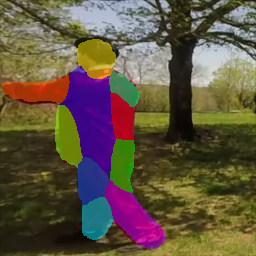} & \includegraphics[width=0.23\columnwidth]{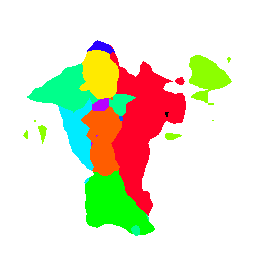}\includegraphics[width=0.23\columnwidth]{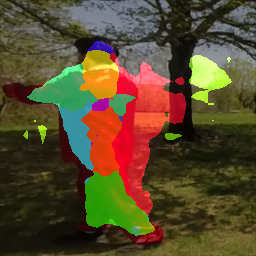} & \includegraphics[width=0.23\columnwidth]{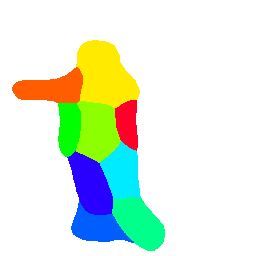}\includegraphics[width=0.23\columnwidth]{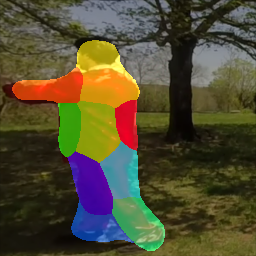}\\
    \end{tabular}
    }
    \captionof{figure}{Additional qualitative co-part segmentation comparisons with recent methods. First column is an input. In next columns, for every method segmentation mask and image with overlayed segmentation are shown.}
    \label{fig:copar-additional}

\end{table*}{}

\section{Implementation details}
\label{sec:impl}
\vspace{-0.3cm}
For a fair comparison, in order to highlight our contributions, we mostly follow the architecture design of FOMM~\cite{fomm}. Similar to FOMM, our region predictor, background motion predictor and pixel-wise flow predictor operate on a quarter of the original resolution, e.g. $64 \times 64$ for $256 \times 256$ images, $96 \times 96$ for $384 \times 384$ and $128 \times 128$ for $512 \times 512$. We use the U-Net~\cite{ronneberger2015u} architecture with five "convolution - batch norm - ReLU - pooling" blocks in the encoder and five "upsample - convolution - batch norm - ReLU" blocks in the decoder for both the region predictor and the pixel-wise flow predictor. For the background motion predictor, we use only five block encoder part. Similarly to FOMM~\cite{fomm}, we use the Johnson architecture~\cite{johnson2016perceptual} for image generation, with two down-sampling blocks, six residual-blocks, and two up-sampling blocks. However, we add skip connections that are warped and weighted by the confidence map. Our method is trained using Adam~\cite{kingma2014adam} optimizer with learning rate $2e-4$ and batch size 48, 20, 12 for $256 \times 256$, $384 \times 384$ and $512 \times 512$ resolutions respectively. During the training process, the networks observe 3M source-driving pairs, each pair selected at random from a random video chunk, and we drop the learning rate by a factor of 10 after 1.8M and 2.7M pairs. We use 4 Nvidia P100 GPUs for training.

The shape-pose disentanglement network consists of 2 identical encoders and 1 decoder. Each encoder consists of 3 "linear - batch norm -ReLU" blocks, with a number of hidden units equal to 256, 512, 1024, and another linear layer with number of units equal to 64. Decoder takes a concatenated input from encoders and applies 3 "linear - batch norm - ReLU" blocks, with sizes 1024, 512, 256. The network is trained on 1M source-driving pairs, organized in batches of 256 images. We use Adam optimizer with learning rate $1e-3$ and drop the learning rate at 660K and 880K pairs.

\section{Background movement}
\label{sec:bg_movement}
While the primary purpose of background modelling is to free up the capacity of the network to better handle the object. For animating articulated objects, background motion is usually unnecessary. Thus, though we estimate background motion, we set it to zero during animation. However, nothing in our framework prevents us from controlling camera motion. Below we show a still background, then move it left, right, and rotate counterclockwise.
\begin{figure}[h]
    \centering
    \includegraphics[width=0.9\linewidth]{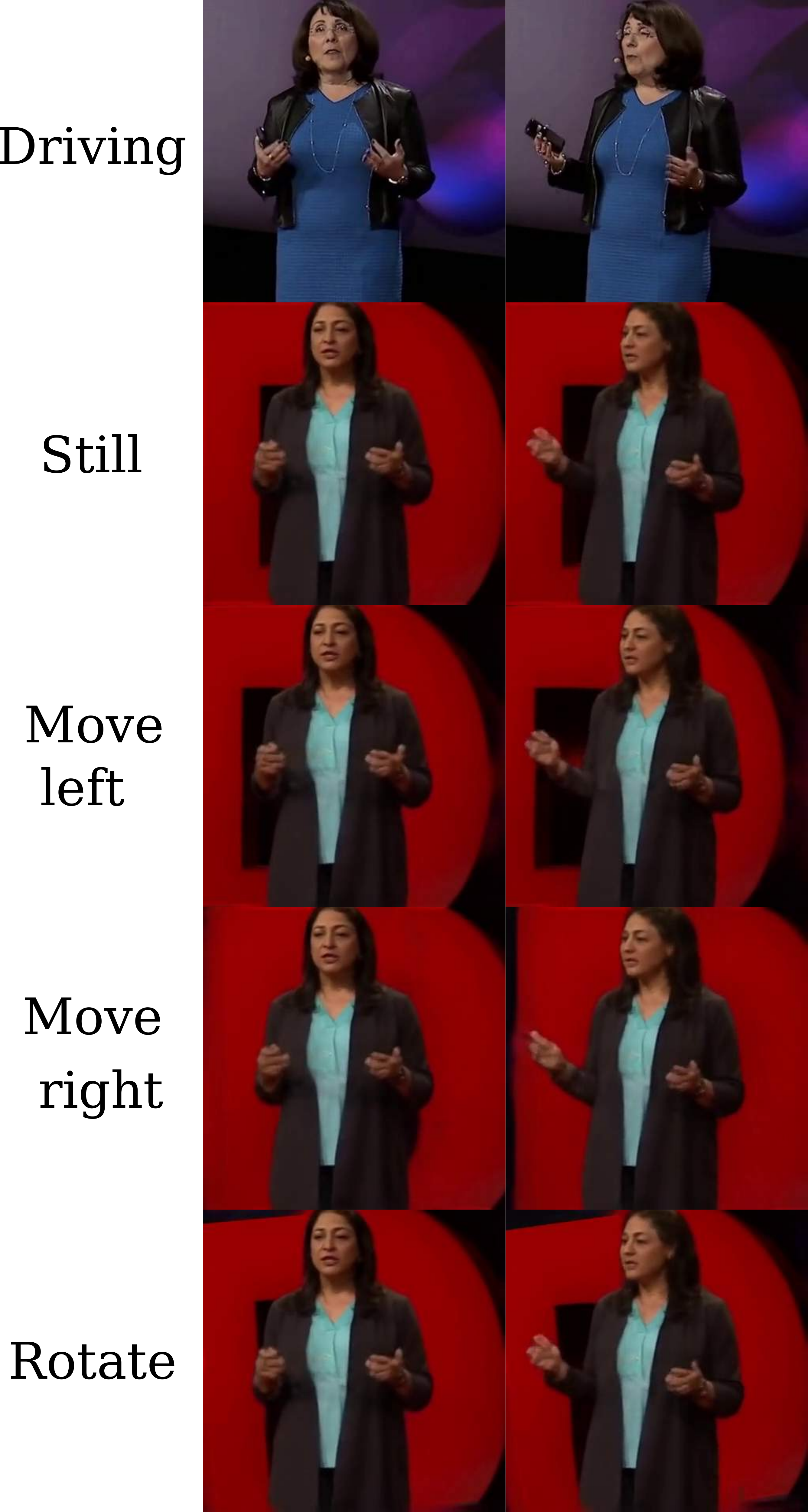}
    \label{fig:bg}
    \caption{Visualizations of background movement. From top to bottom we show driving frame, still background, background that moves left, moves right and rotates counterclockwise.}
\end{figure}

\section{TED-talks dataset creation}
\label{sec:ted}
In order to create the TED-talks dataset, we downloaded 3,035 YouTube videos, shared under the “CC BY – NC – ND 4.0 International” license,\!\!\footnote{This license allows for non-commercial use.} using the "TED talks" query. From these initial candidates, we selected the videos in which the upper part of the person is visible for at least 64 frames, and the height of the person bounding box was at least $384$ pixels. After that, we manually filtered out static videos and videos in which a person is doing something other than presenting. We ended up with 411 videos, and split these videos in 369 training and 42 testing videos. We then split each video into  chunks without significant camera changes (i.e. with no cuts to another camera), and for which the presenter did not move too far from their starting position in the chunk. We cropped the a square region around the presenter, such that they had a consistent scale, and downscaled this region to $384 \times 384$ pixels. Chunks that lacked sufficient resolution to be downscaled, or had a length shorter than 64 frames, were removed. Both the distance moved and the region cropping were achieved using a bounding box estimator for humans~\cite{wu2019detectron2}. We obtained 1,177 training video chunks and 145 test videos chunks.

\end{document}